# Real-time gait planner for human walking using a lower limb exoskeleton and its implementation on Exoped robot

Jafar Kazemi, Sadjaad Ozgoli

*Abstract*—Lower extremity exoskeleton has been developed as a motion assistive technology in recent years. Walking pattern generation is a fundamental topic in the design of these robots. The usual approach with most exoskeletons is to use a pre-recorded pattern as a look-up table. There are some deficiencies with this method, including data storage limitation and poor regulation relating to the walking parameters. Therefore modeling human walking patterns to use in exoskeletons is required. The few existing models provide piece by piece walking patterns, only generating at the beginning of each stride cycle in respect to fixed walking parameters. In this paper, we present a real-time walking pattern generation method which enables changing the walking parameters during the stride. For this purpose, two feedback controlled third order systems are proposed as optimal trajectory planners for generating the trajectory of the x and y components of each joint's position. The boundary conditions of the trajectories are obtained according to some pre-considered walking constraints. In addition, a cost function is intended for each trajectory planner in order to increase the trajectories' smoothness. We use the minimum principle of Pontryagin to design the feedback controller in order to track the boundary conditions in such a way that the cost functions are minimized. Finally, by using inverse kinematics equations, the proper joints angles are generated for and implemented on Exoped robot. The good performance of the gait planner is demonstrated by second derivative continuity of the trajectories being maintained as a result of a simulation, and user satisfaction being determined by experimental testing.

## I. INTRODUCTION

Increase in diseases and accidents related to human mobility, in addition to population ageing, has caused much attention to be given to motion assistive technologies including wearable robots; particularly lower limb exoskeletons. Research on powered human exoskeleton devices dates back to the 1960s in the United States [1] and in the former Yugoslavia [2] for military and medical service purposes respectively [3]. Since then, exoskeleton robots have been well-developed, particularly for medical purposes. Some of them have even hit the market [4].

One of the major challenges in designing these robots is walking pattern generation. There are various methods for planning walking patterns, depending on the application and structure of the exoskeleton [3]. These methods can be classified into three categories: Model-based, sensitivity amplification, and predefined gait trajectory strategies. The model-based strategies take stability into account in order to determine a walking pattern. Stability analysis is mainly based on zero moment point (ZMP) and center of gravity (COG) methods []. This method relies on the accuracy of the human-exoskeleton model and requires various sensors. Sensitivity amplification strategies, including measures like impedance control [5], [6] are applied to walking pattern generation of the exoskeletons with the purpose of load-carrying and rehabilitation. This method is used for situations in which the robot and the wearer have mutual force interaction. Finally, predefined gait trajectory strategies use a pre-recorded pattern of a healthy person as the reference trajectory.

Due to the difficulty of obtaining COG parameters, model-based strategies are not cost-efficient. Furthermore, sensitivity amplification methods are not suitable for the kind of assistance which paraplegic patients need. Therefore predefined gait trajectory methods are the usual approach in most of exoskeletons such as ReWalk [7], eLEGS [8] and ATLAS [9], which is aimed at subjects who are losing their ability to move. There are some shortcomings with this method such as data storage limitation and pattern adjustment depending on different walking parameters and different individuals. Motivated by the above problems, a number of simple models have been developed to generate walking patterns alternated to pre-recorded gaits. The common procedure for these models is to determine the boundary conditions of the joints' trajectory at some specified via-points and to fit a mathematical curve (e.g. polynomial or sinusoidal) to them. In [10] and [11], the hip trajectory is approximated by polynomial and sinusoidal function segments respectively. In [12], two sinusoidal trajectories are proposed as the *x* and *y* components of ankle position. Esfahani *et al.* used polynomial and sinusoidal functions for generating the joint positioning of a biped [13]. Also, Kagawa *et al.* used spline interpolation for joint motion planning of a wearable robot by satisfying determined via-point constraints [14].

In previous works, the trajectories are generated piece by piece as the segments between via-points and the endpoint boundary conditions are fixed within the segments. In other words, the walking patterns are determined at the via-point times. Therefore, changing walking parameters during a stride is not possible. In this paper, we propose a real-time minimum jerk trajectory planner in the joint space for lower limb exoskeletons. In the proposed method, the walking parameters - including step length, maximum foot clearance, and stride time - can be changed during the stride without discontinuity in the second derivative of the trajectories. The idea of our method is to describe the trajectories by a third order system and design a feedback controller to regulate the system's states in order to satisfy the boundary conditions of the trajectory. By introducing (taking) a minimum jerk cost function, the trajectory planning problem is formulated as an optimal control problem with changeable final states. We propose a solution for this problem; using the minimum principle of Pontryagin [15].

It should be noted that in paraplegic exoskeletons, the balance control is executed by the pilot using parallel bars, walkers or crutches. Therefore, the stability of the gait is not taken into account in the walking pattern generation for this kind of exoskeletons.

The remainder of this paper is organized as follows: Section II introduces Exoped robot as our implementation platform. In Section III, the real-time walking pattern generator is proposed by designing a trajectory planner for each x and y component of the position of the joints and determining the boundary conditions of the trajectories. Our simulation and experimental results are presented in Section IV. Finally, we conclude the paper in Section V.

## II. EXOPED

Exoped has 4 DOFs (2 DOFs on each leg: 1 at the hip and 1 at the knee), driven by 4 brushless electronically communicated (EC) motors. The motors from the Maxon model "EC 90 flat", fed by 36V, are used. Each motor is coupled with a 1:135 gearhead and internal hall sensors are used to indicate the position. Stm32f429 and PID controller are employed as high level and low level controllers, respectively. The forward kinematics of Exoped can be described as follows:

$$X_R = l_T \times \sin(\theta_{Rh}) + l_S \times \sin(\theta_{Rh} + \theta_{Rk})$$
$$Y_R = -l_T \times \cos(\theta_{Rh}) - l_S \times \cos(\theta_{Rh} + \theta_{Rk})$$
$$X_L = l_T \times \sin(\theta_{Lh}) + l_S \times \sin(\theta_{Lh} + \theta_{Lk})$$
$$Y_L = -l_T \times \cos(\theta_{Lh}) - l_S \times \cos(\theta_{Lh} + \theta_{Lk}) \quad (1)$$

where R and L refer to right and left leg, respectively. $\theta_{*h}$ and $\theta_{*k}$ denote the angles of the hip and knee joint of each leg. $l_T$ and $l_S$ represent the length of thigh and shin respectively and $X_R, Y_R, X_L$ and $Y_L$ are expressed by:

$$X_R = X_{Ra} - X_h$$
$$Y_R = Y_{Ra} - Y_h$$
$$X_L = X_{La} - X_h$$
$$Y_L = Y_{La} - Y_h \quad (2)$$

in which $\vec{r}_h = (X_h, Y_h)$, $\vec{r}_{Ra} = (X_{Ra}, Y_{Ra})$, and $\vec{r}_{La} = (X_{La}, Y_{La})$ represent the position of hip, right ankle and left ankle in sagittal plan, respectively. The initial position of the right ankle is defined as the origin point of coordination. The defined parameters are shown in Fig. 1.

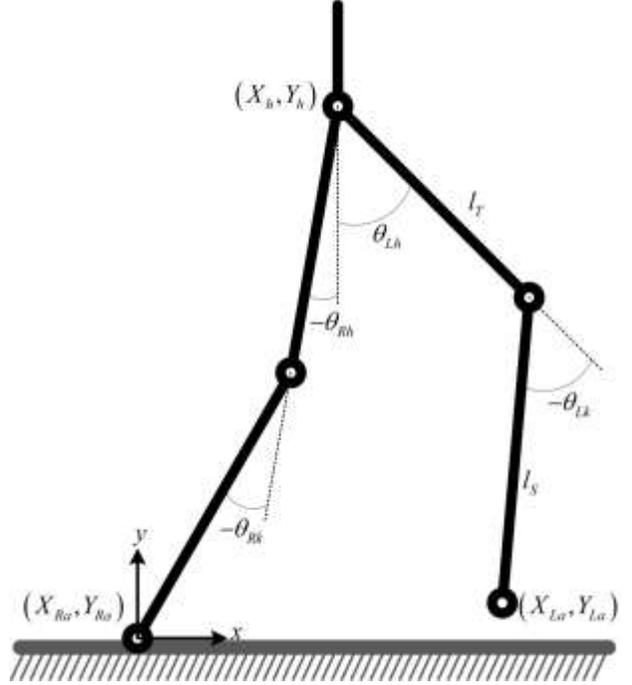

Fig. 1. Robot parameter description.

The inverse kinematic equations of the robot are described as the following:

$$\theta_{Lk} = -\cos^{-1}\left(\frac{X_L^2 + Y_L^2 + l_T^2 - l_S^2}{2 \times l_T \sqrt{X_L^2 + Y_L^2}}\right)$$
$$- \cos^{-1}\left(\frac{X_L^2 + Y_L^2 + l_S^2 - l_T^2}{2 \times l_S \sqrt{X_L^2 + Y_L^2}}\right)$$
$$\theta_{Lh} = \tan^{-1}\left(\frac{-X_L}{Y_L}\right) - \cos^{-1}\left(\frac{X_L^2 + Y_L^2 + l_T^2 - l_S^2}{2 \times l_T \sqrt{X_L^2 + Y_L^2}}\right)$$
$$\theta_{Rk} = -\cos^{-1}\left(\frac{X_R^2 + Y_R^2 + l_T^2 - l_S^2}{2 \times l_T \sqrt{X_R^2 + Y_R^2}}\right)$$
$$- \cos^{-1}\left(\frac{X_R^2 + Y_R^2 + l_S^2 - l_T^2}{2 \times l_S \sqrt{X_R^2 + Y_R^2}}\right)$$
$$\theta_{Rh} = \tan^{-1}\left(\frac{-X_R}{Y_R}\right) - \cos^{-1}\left(\frac{X_R^2 + Y_R^2 + l_T^2 - l_S^2}{2 \times l_T \sqrt{X_R^2 + Y_R^2}}\right) \quad (3)$$

## III. REAL-TIME WALKING PATTERN GENERATION

Fig. 2 shows the overall control schematic of the robot. As shown, a well-defined algorithm calculates the walking parameters according to the stability of the robot and particular conditions; e.g. patient's dimension, environment conditions etc. The walking parameters are used as the inputs for a pattern generator block. These parameters are step length, maximum foot clearance, and step time interval denoted by $L_s$, $H_s$ and $t_s$, respectively. The pattern generator provides the appropriate position of the joints in a sagittal plane and subsequently, the positions are transformed into joint angles using the inverse kinematic equations. Finally, a PID feedback controller is employed to regulate the joint angles.

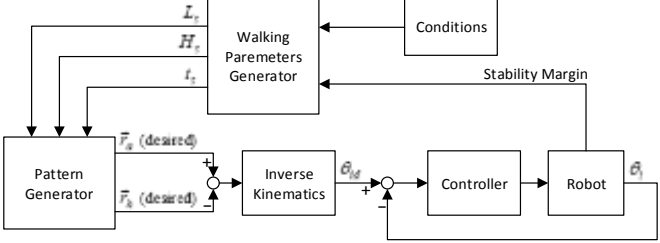

Fig. 2. Exoskeleton control block diagram.

The pattern generator consists of custom trajectory planners that generate the trajectory of the $x$ and $y$ components of each joint's position. Continuity, smoothness, and taking walking constraints into account are the main objectives to be considered in trajectory planning. In most previous works on gait planning, the walking parameters are considered as constant during the stride. The most common method of trajectory planning is fitting a mathematical curve (e.g. polynomial or sinusoidal) to some boundary conditions obtained by walking parameters. From a systems point of view, this kind of trajectory planner can be described as an input-output zero order system (Fig 3). With this description, a discontinuity of the inputs caused by a change in walking parameters yields to a discontinuity in the output.

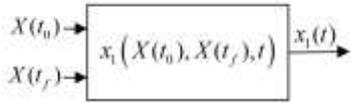

Fig. 3. The most common trajectory planner. $X(t_0) = [x_1(t_0), \dot{x}_1(t_0), \ddot{x}_1(t_0)]$ and $X(t_f) = [x_1(t_f), \dot{x}_1(t_f), \ddot{x}_1(t_f)]$ refer to the boundary conditions of the trajectory.

By using the control scheme depicted in Fig. 2, the walking parameters may be updated at any time due to a change in the stability of the robot or the particular conditions. Accordingly, an optimal trajectory planner is required to adjust the trajectory according to the change in the parameters in order to maintain the second derivative continuity. From a systems point of view, maintaining second derivative continuity of the output (trajectory) against the discontinuity of the input (boundary conditions) require a third order system. For this purpose, we propose feedback controlled third order systems as the optimal trajectory planners for generating the trajectory of the joints' $x$ and $y$ component, as follows.

*A. The x component of the joints*

Fig. 4 shows a general trajectory shape for the $x$ component of the position of the joints, starting from initial condition $X(t_0) = [x_1(t_0) \ \dot{x}_1(t_0) \ \ddot{x}_1(t_0)]^T$ converging to the final value $X(t_f) = [x_1(t_f) \ \dot{x}_1(t_f) \ \ddot{x}_1(t_f)]^T$, with the minimum curvature.

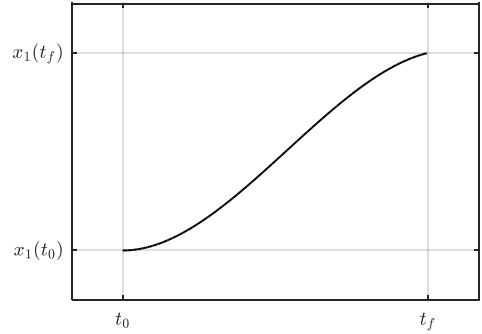

Fig. 4. The general trajectory shape for the x component of the position of the joints.

In order to generate this type of trajectory, we propose the feedback controlled third order system depicted in Fig. 5, which can be formulated as follows:

$$S_x : \begin{cases} \dot{x}_1 = x_2 \\ \dot{x}_2 = x_3 \\ \dot{x}_3 = u \end{cases} \qquad (4)$$

It is obvious that a finite $u$ yields a continuous trajectory with a continuous second derivative. In addition, the cost function denoted by $J_x$ is intended to be minimized in order to increase the smoothness of the trajectory.

$$J_x = \int_{t_0}^{t_f} \left(\frac{d^3 x_1}{dt^3}\right)^2 dt \qquad (5)$$

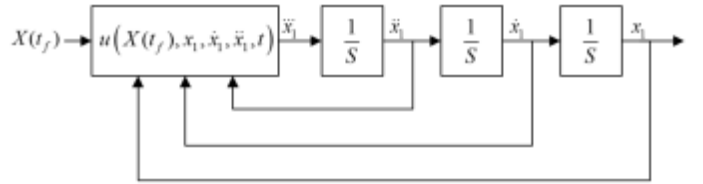

Fig. 5. The proposed trajectory planner for the x component of the position of the joints. $X(t_f) = [x_1(t_f), \dot{x}_1(t_f), \ddot{x}_1(t_f)]$ refers to the final boundary conditions of the trajectory.

As the result, the optimal trajectory planner can be designed by calculating the feedback control low $u$ to move the state of the system $S_x$ to the changeable final condition $X(t_f)$ in such a way that the cost function $J_x$ is minimized. We have presented Theorem 1 to solve this optimal control problem. Lemma 1 is used to prove Theorem 1.

**Lemma 1:** $u(t) = v_x(t)$ steers the states of the system $S_x$ from the initial value $X(t_0) = [x_1(t_0) \ x_2(t_0) \ x_3(t_0)]^T$ to $X(t_f) = [x_1(t_f) \ x_2(t_f) \ x_3(t_f)]^T$ final value in such a way that the cost function $J_x$ has been minimized. $v_x(t)$ is obtained as the following:

$$v_x(t) = a(t - t_0)^2 - b(t - t_0) + c \qquad (6)$$

where

$$a = \frac{360(x_1(t_f)-x_1(t_0))}{(t_f-t_0)^5} - \frac{180(x_2(t_f)+x_2(t_0))}{(t_f-t_0)^4} + \frac{30(x_3(t_f)-x_3(t_0))}{(t_f-t_0)^3}$$

$$b = \frac{360(x_1(t_f)-x_1(t_0))}{(t_f-t_0)^4} - \frac{24(7x_2(t_f)+8x_2(t_0))}{(t_f-t_0)^3} + \frac{12(2x_3(t_f)-3x_3(t_0))}{(t_f-t_0)^2}$$

$$c = \frac{60(x_1(t_f)-x_1(t_0))}{(t_f-t_0)^3} - \frac{12(2x_2(t_f)+3x_2(t_0))}{(t_f-t_0)^2} + \frac{3(x_3(t_f)-3x_3(t_0))}{(t_f-t_0)}$$

**Proof**:

According to the minimum principle of Pontryagin [1], minimization of $J_x$ can be achieved by minimizing the *Hamiltonian* function defined as

$$H(X(t), v_x(t), P(t)) = v_x^2(t) + p_1(t)x_2(t) + p_2(t)x_3(t) + p_3(t)v_x(t) \qquad (7)$$

where $P(t)$ is defined as co-state vector. The optimal trajectories $X^*(t)$ and $v_x^*(t)$ can be achieved by satisfying the following conditions:

$$\dot{X}^*(t) = \frac{\partial H}{\partial P}(X^*(t), v_x^*(t), P^*(t), t)$$
$$\dot{P}^*(t) = -\frac{\partial H}{\partial X}(X^*(t), v_x^*(t), P^*(t), t)$$
$$0 = \frac{\partial H}{\partial v_x}(X^*(t), v_x^*(t), P^*(t), t) \qquad (8)$$

where the symbol * refers to the extremals of $X(t)$, $v_x(t)$ and $P(t)$. The necessary conditions for optimality can be written as

$$\dot{x}_1^*(t) = x_2^*(t)$$
$$\dot{x}_2^*(t) = x_3^*(t)$$
$$\dot{x}_3^*(t) = -0.5 p_3^*(t)$$
$$\dot{p}_1^*(t) = 0 \qquad (9)$$
$$\dot{p}_2^*(t) = -p_1^*(t)$$
$$\dot{p}_3^*(t) = -p_2^*(t)$$
$$v_x^*(t) = -0.5 p_3^*(t).$$

By applying the initial condition $X(t_0)$ and the final value $X(t_f)$, the optimal control function obtained for $t \in [t_0, t_f]$ is (6).

Lemma 1 represents an open-loop control method to steer the state of system $S_x$ from a specific initial condition to a fixed final value along with minimizing $J_x$. This control method is not robust against disturbance due to the nature of the open-loop control methods. In addition, $v_x(t)$ is determined in $t = t_0$ for $t \in [t_0, t_f]$ interval and the final value is not changeable in $t \in (t_0, t_f]$. Using a closed-loop control method and online calculation of $u$ brings about disturbance rejection and makes the final value changeable. For this purpose, Theorem 1 is represented.

**Theorem 1:** Feedback control law $u_x(t)$ steers the states of the system $S_x$ from any initial value to $X(t_f) = [x_f \ \dot{x}_f \ \ddot{x}_f]^T$ final value, in such a way that the cost function $J_x$ is minimized. $u_x(t)$ is obtained for $t < t_f$ as the following:

$$u_x(t) = \frac{60(x_f - x_1(t))}{(t_f - t)^3} - \frac{12(2\dot{x}_f + 3x_2(t))}{(t_f - t)^2} + \frac{3(\ddot{x}_f - 3x_3(t))}{(t_f - t)} \qquad (10)$$

**Proof**:

Considering state feedback $X(t) = [x_1(t) \ x_2(t) \ x_3(t)]^T$ as the initial condition of the system $S_x$ at any moment leads to disturbance rejection. On the other hand, online calculation of $u$ in respect to the new defined initial condition allows the final value $X(t_f)$ to be changeable. Considering $X(t)$ as $X(t_0)$ corresponds to putting $t$ instead of $t_0$ in the formulation of $v_x(t)$ in Lemma 1. In other words:

$$u_x(t) = \{v_x(t) | t_0 = t\}. \qquad (11)$$

Therefore, from Lemma 1 and (11), $u_x(t)$ is obtained as (10).

The trajectory obtained by applying $u_x(t)$ to system $S_x$ is denoted by $T_x(t, t_f, x_f, \dot{x}_f, \ddot{x}_f)$, where:

$$T_x(t, t_f, x_f, \dot{x}_f, \ddot{x}_f) = \{x_1(t) | u = u_x(t), t < t_f\} \qquad (12)$$

To achieve a unique result, the initial states are assumed to be zero.

### B. The y component of the joints

Fig. 6 shows a general trajectory shape for the $y$ component of the position of the joints, rising from the initial condition $Y(t_0) = [y_1(t_0) \ \dot{y}_1(t_0) \ \ddot{y}_1(t_0)]^T$ to a peak of $y_p$, and converging to the final value $Y(t_f) = [y_1(t_f) \ 0 \ 0]^T$, with the minimum curvature.

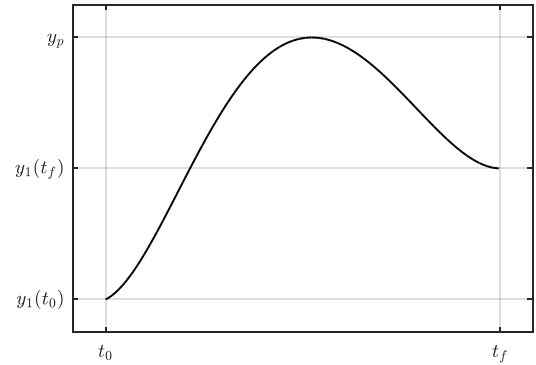

Fig. 6. The general trajectory shape for the y component of the position of the joints.

We propose the feedback controlled third order system depicted in Fig. 7 in order to plan this trajectory, which can be formulated as follows:

$$S_y: \begin{cases} \dot{y}_1 = y_2 \\ \dot{y}_2 = y_3 \\ \dot{y}_3 = u \end{cases} \qquad (13)$$

In order to increase the trajectory's smoothness, the cost function denoted by $J_y$ is intended as (14). Minimizing $J_y$ causes a peak on the trajectory in addition to the smoothness increment, wherein parameter $k$ determines the value of the peak.

$$J_y = \int_{t_0}^{t_f} \left(\frac{d^3 y_1}{dt^3}\right)^2 + k y_1(t) \ dt \qquad (14)$$

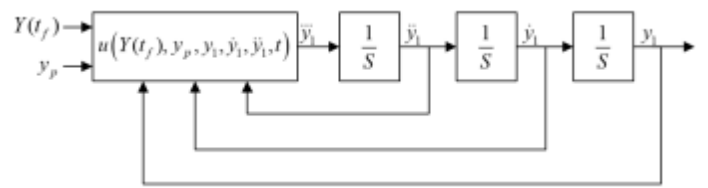

Fig. 7. The proposed trajectory planner for the y component of the position of the joints. $Y(t_f) = [y_1(t_f), 0, 0]$ refers to the final boundary conditions of the trajectory.

We have presented Theorem 2 to design the proper feedback control low $u$ in order to move the states of the system $S_y$ to the final condition $Y(t_f)$, along with minimizing the cost function $J_y$. Lemma 2 is used to prove of Theorem 2.

**Lemma 2:** $u(t) = v_y(t)$ steers the states of the system $S_y$ from the initial value $Y(t_0) = [y_1(t_0) \ y_2(t_0) \ y_3(t_0)]^T$ to $Y(t_f) = [y_1(t_f) \ 0 \ 0]^T$ final value in such a way that the cost function $J_y$ is minimized. $v_y(t)$ is obtained as the following:

$$v_y(t) = \frac{k}{12}(t-t_0)^3 - a(t-t_0)^2 + b(t-t_0) - c \quad (15)$$

where

$$a = \frac{360(y_1(t_0)-y_1(t_f))}{(t_f-t_0)^5} + \frac{180 y_2(t_0)}{(t_f-t_0)^4} + \frac{30 y_3(t_0)}{(t_f-t_0)^3} + \frac{k(t_f-t_0)}{8}$$

$$b = \frac{360(y_1(t_0)-y_1(t_f))}{(t_f-t_0)^4} + \frac{192 y_2(t_0)}{(t_f-t_0)^3} + \frac{36 y_3(t_0)}{(t_f-t_0)^2} + \frac{k(t_f-t_0)^2}{20}$$

$$c = \frac{60(y_1(t_0)-y_1(t_f))}{(t_f-t_0)^3} + \frac{36 y_2(t_0)}{(t_f-t_0)^2} + \frac{9 y_3(t_0)}{(t_f-t_0)} + \frac{k(t_f-t_0)^3}{240}$$

**Proof**:

As in the proof of Lemma 1, the *Hamiltonian* function defined as

$$H(Y(t), v_y(t), P(t)) = v_y^2(t) + k y_1(t) + p_1(t) y_2(t) + p_2(t) y_3(t) + p_3(t) v_y(t). \quad (16)$$

The necessary conditions for optimality can be written as
$$\dot{y}_1^*(t) = y_2^*(t)$$
$$\dot{y}_2^*(t) = y_3^*(t)$$
$$\dot{y}_3^*(t) = -0.5 p_3^*(t)$$
$$\dot{p}_1^*(t) = -k \quad (17)$$
$$\dot{p}_2^*(t) = -p_1^*(t)$$
$$\dot{p}_3^*(t) = -p_2^*(t)$$
$$v_y^*(t) = -0.5 p_3^*(t).$$

By applying the initial condition $Y(t_0)$ and the final value $Y(t_f)$, the optimal control function can be obtained for $t \in [t_0, t_f]$ as (15).

**Theorem 2**: Feedback control law $u_y(t)$ steers the states of the system $S_y$ from any initial value to $Y(t_f) = [y_f \ 0 \ 0]^T$ final value, in such a way that the cost function $J_y$ is minimized. $u_y(t)$ is obtained as the following:

$$u_y(t) = \frac{60(y_f - y_1(t))}{(t_f - t)^3} - \frac{36 y_2(t)}{(t_f - t)^2} - \frac{9 y_3(t)}{(t_f - t)} - \frac{K(t_f - t)^3}{240} \quad (18)$$

**Proof**:

As in the proof of Theorem 1:
$$u_y(t) = \{v_y(t) | t_0 = t\}, \quad (19)$$

From Lemma 2 and (19), $u_y(t)$ is obtained as (18).

**Calculation of $k$:**

The proper value of $k$ for generating a trajectory with a peak of $y_p$ can be obtained as

$$k^* = -\frac{5 \times 8!}{(t_f - t_0)^6} \times \frac{(y_p - y_0)(y_p - y_f)}{(y_p - y_0) + (y_p - y_f)} \quad (20)$$

**Proof**:

According to the designed feedback control law (18) we have

$$\int_{t_0}^{t_f} y_1(t) dt \bigg|_{k=k^*} - \int_{t_0}^{t_f} y_1(t) dt \bigg|_{k=0} = -k^* \frac{(t_f - t_0)^7}{5 \ast 8!} \quad (21)$$

Regarding to Fig. 8, the approximate integral of $y_1(t)$ for $k = 0$ (Left) and for $k = k^*$ (Right) can be calculated numerically as

$$\int_{t_0}^{t_f} y_1(t) dt \bigg|_{k=0} \cong \frac{y_0 + y_f}{2}(t_f - t_0) \quad (22)$$

$$\int_{t_0}^{t_f} y_1(t) dt \bigg|_{k=k^*} \cong \frac{y_0 + y_p}{2}(t_p - t_0) + \frac{y_p + y_f}{2}(t_f - t_p) \cong \frac{2 y_p^2 - y_0^2 - y_f^2}{4 y_p - 2 y_0 - 2 y_f}(t_f - t_0) \quad (23)$$

Where the parameter $t_p$ denoted in Fig. 8 (Right) refers to the peak time and is obtained approximately by

$$t_p \cong \frac{y_p - y_0}{2 y_p - y_0 - y_f} \quad (24)$$

Using (21), (22), and (23) the proper value of $k$ can extracted as (20).

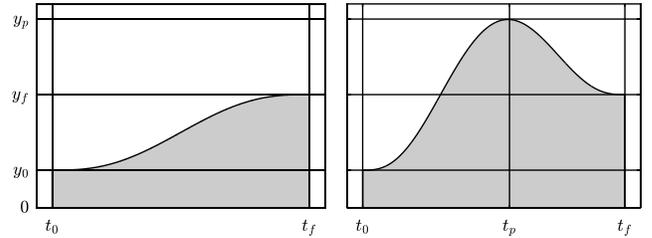

Fig. 8. The integral of $y_1(t)$ for $k = 0$ (Left) and $k = k^*$ (Right).

The trajectory obtained by applying $u_y(t)$ to the system $S_y$ is denoted by $T_y(t, t_0, t_f, y_0, y_p, y_f)$, where:

$$T_y(t, t_0, t_f, y_0, y_p, y_f) = \{y_1(t) | u = u_y(t), t_0 \le t < t_f\} \quad (20)$$

To achieve a unique result, the initial states are assumed as zero.

$T_x$ and $T_y$ are employed as optimal smooth trajectory planners for planning the trajectory of $x$ and $y$ component of the position of each joint respectively. The only parameters needed for planning the trajectories are $(t_f, x_f, \dot{x}_f, \ddot{x}_f)$ for $T_x$, and $(t_0, t_f, y_0, y_p, y_f)$ for $T_y$, which can be extracted as the endpoint boundary conditions of the trajectories.

In section B, the endpoint boundary conditions of the trajectory of each joint are calculated in respect to stride parameters.

C. *Walking pattern generator*

Fig. 9 demonstrates the fundamental parameters involved in a stride cycle beginning from the initial ground contact position denoted by $X_{Ra0}$ and $X_{La0}$. In this stride, the right leg (Red) is in the swing phase while the left one (Black) stays in the stance phase. The robot's posture is displayed in three instants of time demonstrating the walking constraints, which is given as

$$\theta_{Lk}(t) = \theta_{Rk}(t) = 0 \text{ for } t = t_0, t_0 + t_s$$

$$\max(Y_h) = l$$
$$\max(Y_{Ra}) = H_s \tag{21}$$

where $l = l_S + l_T$. The parameters $\Delta_{h0}$ and $\Delta_h$ defined in Fig. 9 can be calculated as

$$\Delta_{h0} = l - \sqrt{l^2 - \left(\frac{X_{La0} - X_{Ra0}}{2}\right)^2}$$
$$\Delta_h = l - \sqrt{l^2 - \left(\frac{L_s}{4}\right)^2} \tag{22}$$

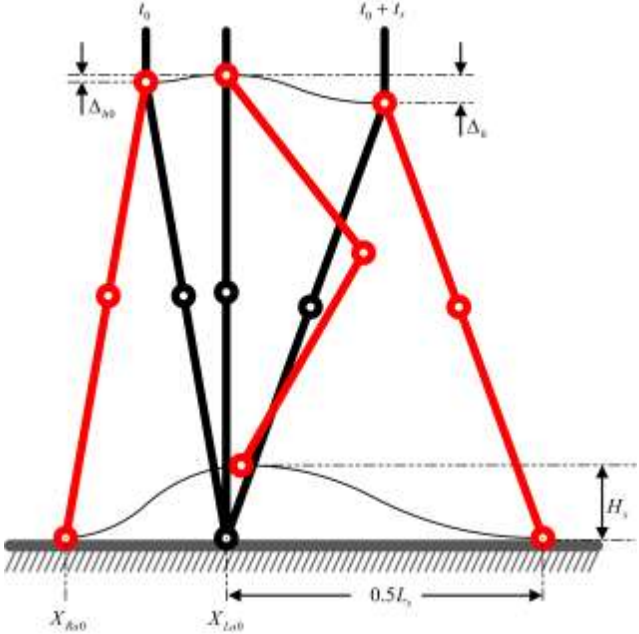

Fig. 9. Parameter description in a stride cycle.

Boundary conditions of the $x$ and $y$ components of each joint's position can be obtained as shown in Table 1 and Table 2 respectively.

TABLE 1: BOUNDARY CONDITION OF THE X COMPONENT OF THE JOINTS POSITION.

| $X_*$ | $x_f = X_*(t_0 + t_s)$ | $\dot{x}_f = \dot{X}_*(t_0 + t_s)$ | $\ddot{x}_f = \ddot{X}_*(t_0 + t_s)$ |
|---|---|---|---|
| $X_h$ | $X_{La0} + 0.25L_s$ | 0 | 0 |
| $X_{La}$ | $X_{La0}$ | 0 | 0 |
| $X_{Ra}$ | $X_{La0} + 0.5L_s$ | 0 | 0 |

TABLE 2: BOUNDARY CONDITION OF THE Y COMPONENT OF EACH JOINTS POSITION.

| $Y_*$ | $y_0 = Y_*(t_0)$ | $y_f = Y_*(t_0 + t_s)$ | $y_p = max(Y_*)$ |
|---|---|---|---|
| $Y_h$ | $l - \Delta_{h0}$ | $l - \Delta_h$ | $l$ |
| $Y_{La}$ | 0 | 0 | 0 |
| $Y_{Ra}$ | 0 | 0 | $H_s$ |

Using the proposed optimal trajectory planners $T_x$ and $T_y$, and the given endpoint boundary conditions, a real-time walking pattern generation in the joint space is developed as

$$X_{La}(t) = T_x(t, t_s, X_{La0}, 0, 0)$$
$$Y_{La}(t) = T_y(t, t_0, t_s, 0, 0, 0)$$
$$X_{Ra}(t) = T_x(t, t_s, X_{La0} + 0.5L_s, 0, 0)$$
$$Y_{Ra}(t) = T_y(t, t_0, t_s, 0, H_s, 0)$$
$$X_h(t) = T_x(t, t_s, X_{La0} + 0.25L_s, 0, 0)$$
$$Y_h(t) = T_y(t, t_0, t_s, l - \Delta_{h0}, l, l - \Delta_h)$$
$$t_0 \leq t < t_f \tag{23}$$

By calculating $X_R$, $Y_R$, $X_L$ and $Y_L$ from (2) and applying inverse kinematics given by (3), the joint angles will be obtained.

IV. EXPERIMENTAL RESULTS AND DISCUSSION

A. Simulation of the designed trajectory planners

The designed trajectory planners $T_x$ and $T_y$ play the main roles in the walking pattern generation and have an effect on the performance of the gait. Therefore a performance analysis of the trajectory planners is required, especially for their response to the changing boundary conditions.

Resulting from the simulation, Fig. 10 and Fig. 11 show the trajectories planned by $T_x$, for fixed and changeable boundary conditions respectively. Fig. 10 shows the generated trajectory starting from the initial condition $[x(0) \; \dot{x}(0) \; \ddot{x}(0)] = [0 \; 0 \; 0]$ and ending up at the endpoint boundary condition $[x_f \; \dot{x}_f \; \ddot{x}_f] = [2 \; 1 \; 1]$ at $t_f = 5$. While in Fig. 11, the boundary conditions change from $[x_f \; \dot{x}_f \; \ddot{x}_f] = [2 \; 1 \; 1]$ to $[x_f \; \dot{x}_f \; \ddot{x}_f] = [1 \; -0.5 \; -1]$ at $t = 2$ and the boundary time is brought forward from $t_f = 5$ to $t_f = 4$ at $t = 3$. Similarly for $T_y$, Fig. 12 and Fig. 13 represent the trajectories, respectively for fixed and changeable boundary conditions. Fig. 12 shows the trajectory generated by $T_y$ for $t_f = 5$, $y_p = 2$, $y_f = 1$ and the initial condition $[y(0) \; \dot{y}(0) \; \ddot{y}(0)] = [0 \; 0 \; 0]$. While in Fig. 13, the boundary parameters change three times at $t = 1, 2,$ and 3. The accurate tracking of the boundary conditions and maintaining second derivative continuity against the change of parameters illustrates the good performance of the trajectory planners.

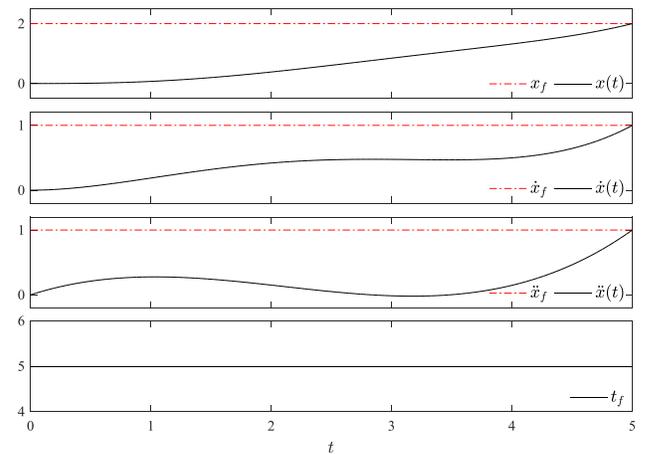

Fig. 10. Trajectory planned by $T_x$ for fixed boundary conditions.

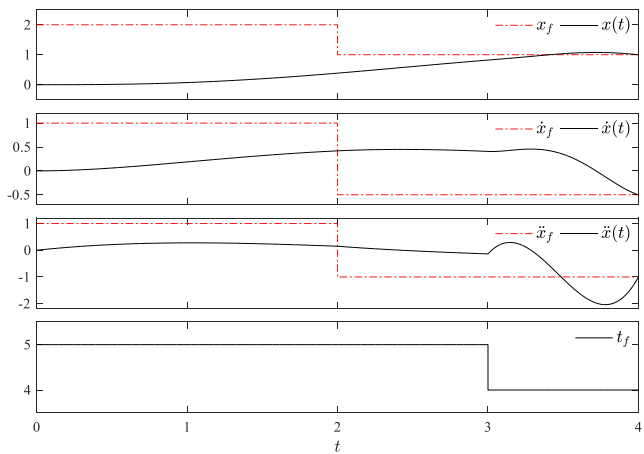

Fig. 11. Trajectory planned by $T_x$ for changeable boundary conditions.

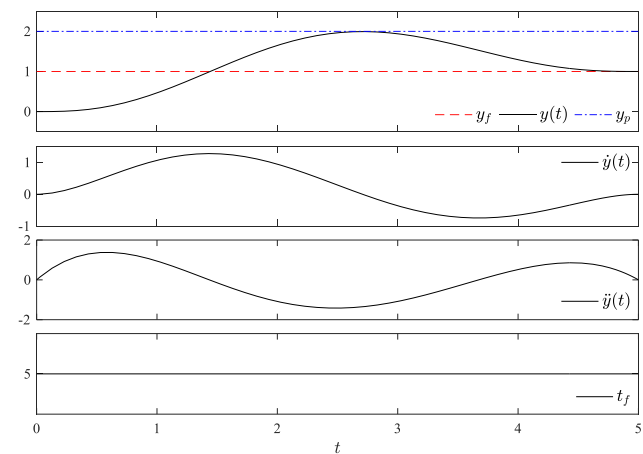

Fig. 12. Trajectory planned by $T_y$ for fixed boundary conditions.

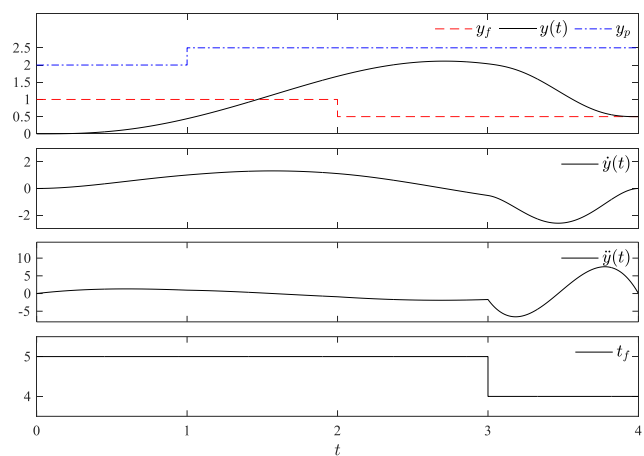

Fig. 13. Trajectory planned by $T_y$ for changeable boundary conditions.

## B. The real-time walking pattern implementation

For evaluating the proposed walking pattern generator, two experiments were carried out with different walking parameters. In the first experiment, walking begins from a standing pose and the right leg takes the first half step. After taking two strides sequentially with the left and right legs, the walking ends with a half step taken by the left leg. Fig. 14 shows the walking parameters related to the first experiment, which took 12 seconds. As shown, the half-step and full-step time was 2 and 4 seconds respectively. Maximum foot clearance ($H_s$) and step length ($L_s$) were considered as 10 and 60 cm respectively. The zero step length from $t = 10s$ to $t = 12s$ corresponds to the final half-step.

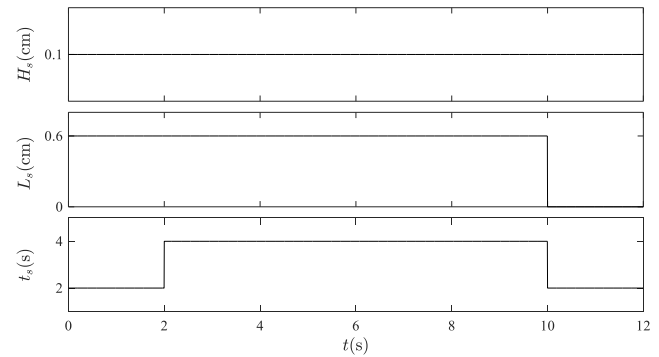

Fig. 14. Walking parameters of the first experiment.

Figs. 15 and 16 show the real-time position of the joints generated by the proposed method for the first experiment. As shown by the figures, the required walking parameters were satisfied. Moreover, all of the trajectories had a continuous second derivative.

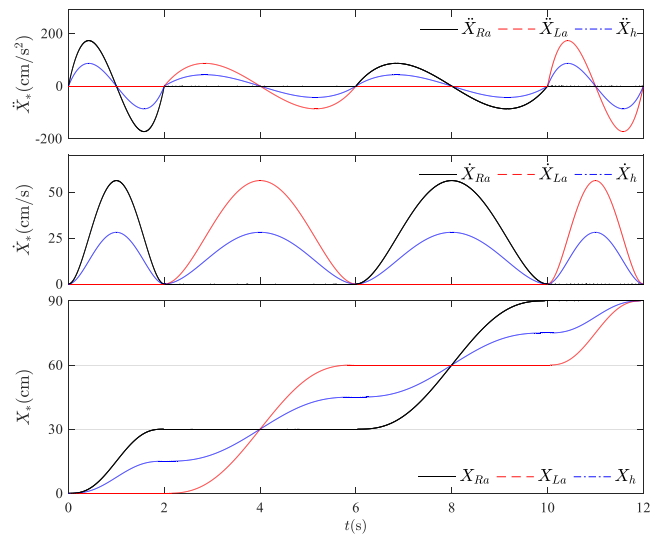

Fig. 15. The x component of the position of the joints in the first experiment.

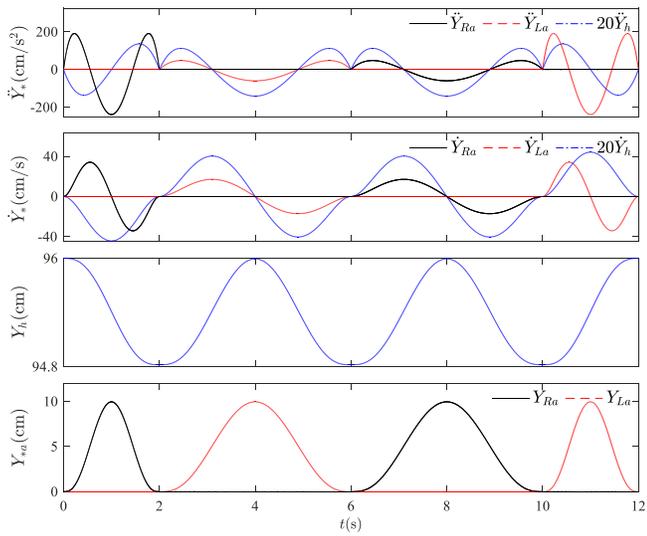

Fig. 16. The y component of the position of the joints in the first experiment.

By applying inverse kinematic transformations, the desired angle trajectories of the joints were obtained in real-time as shown in Figs. 17-20. A PID controller was used for regulation of each motor's reference input. The obtained reference angle of the joints compared to the real angles measured by the hall sensors are depicted in the figures. Also, the control signal and the motor currents corresponding to each joint are shown.

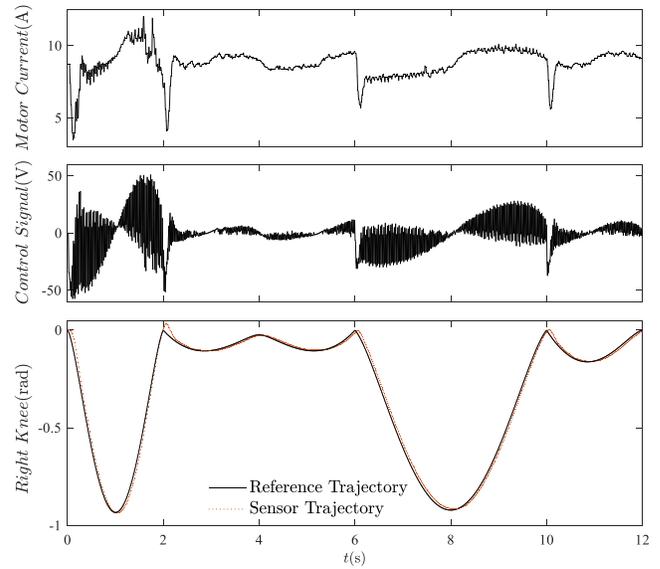

Fig. 18. The reference angle, the real angle, the control signal, and the motor current of the right knee.

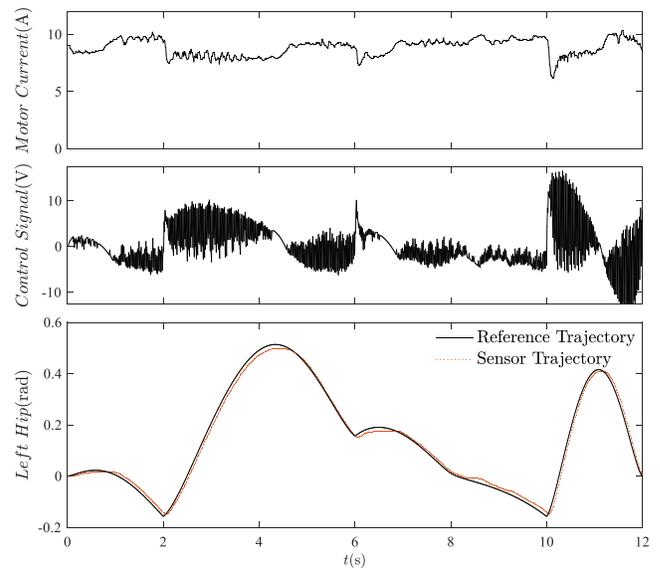

Fig. 19. The reference angle, the real angle, the control signal, and the motor current of the left hip.

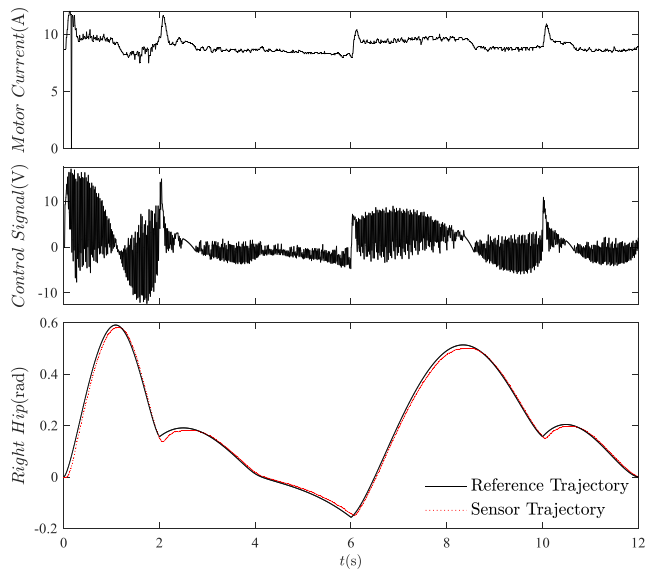

Fig. 17. The reference angle, the real angle, the control signal, and the motor current of the right hip.

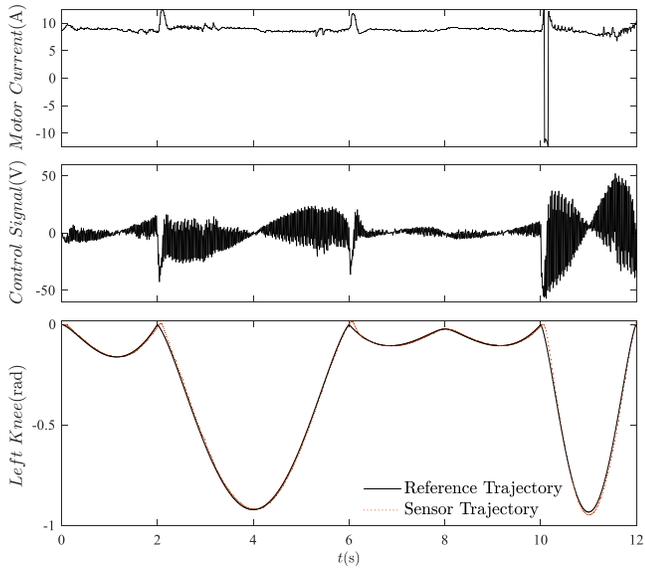

Fig. 20. The reference angle, the real angle, the control signal, and the motor current of the left knee.

The second experiment was carried out to evaluate the efficacy of the proposed method in response to changes in the walking parameters. Fig. 21 shows the walking parameters during the experiment. As shown, maximum foot clearance and step length increased in $t = 1.8s$ and $t = 6s$ respectively. Also, walking speed decreased in $t = 4.5s$. The walking ends with a decision to put the left leg down $20cm$ ahead of the right leg in 1.2 seconds. This corresponds to change $L_s$ to $0.4m$ and $t_s$ to $2.7s$.

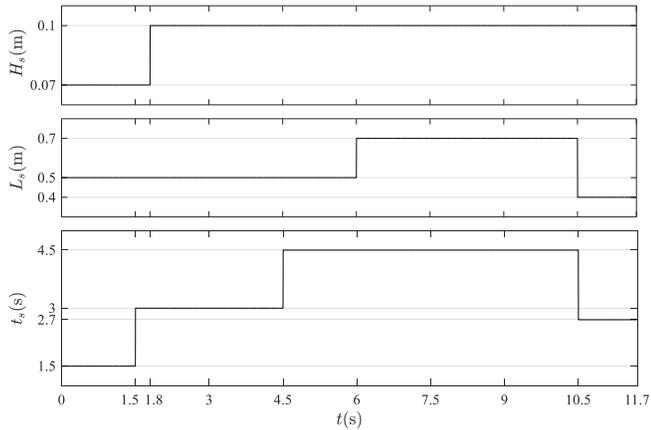

Fig. 21. Walking parameters of the first experiment.

Figs. 22 and 23 show the generated real-time position of the joints. As shown by the figures, the required walking parameters were satisfied in addition to maintaining continuity of the second derivatives. Figs. 24-27 show the implementation results of the second experiment.

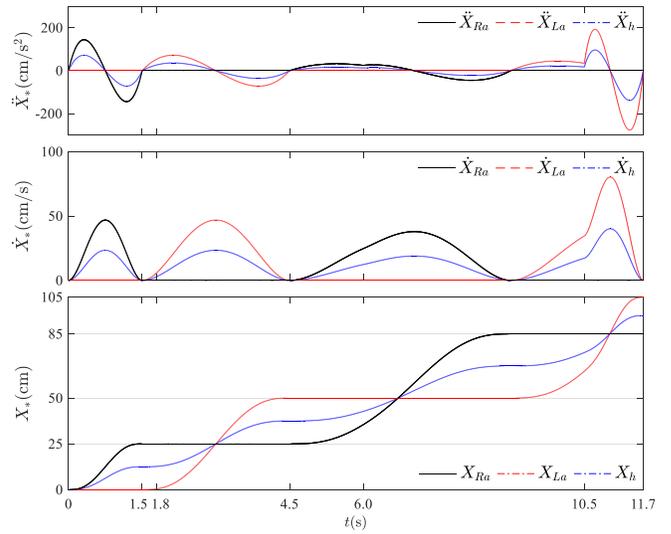

Fig. 22. The x component of the position of the joints in the second experiment.

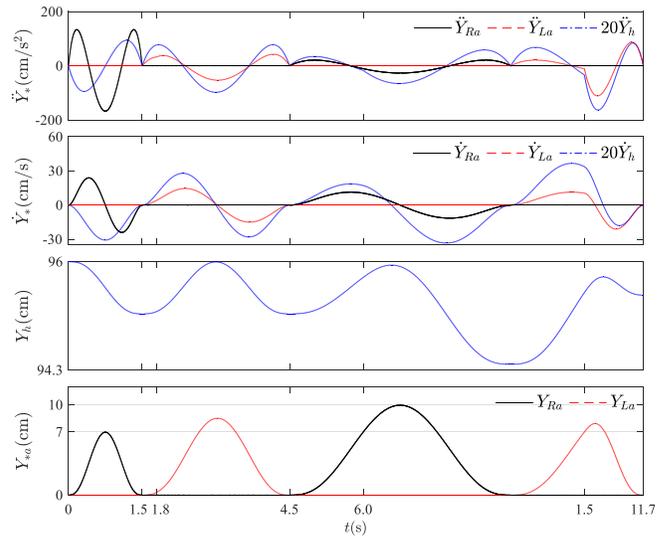

Fig. 23. The y component of the position of the joints in the second experiment.

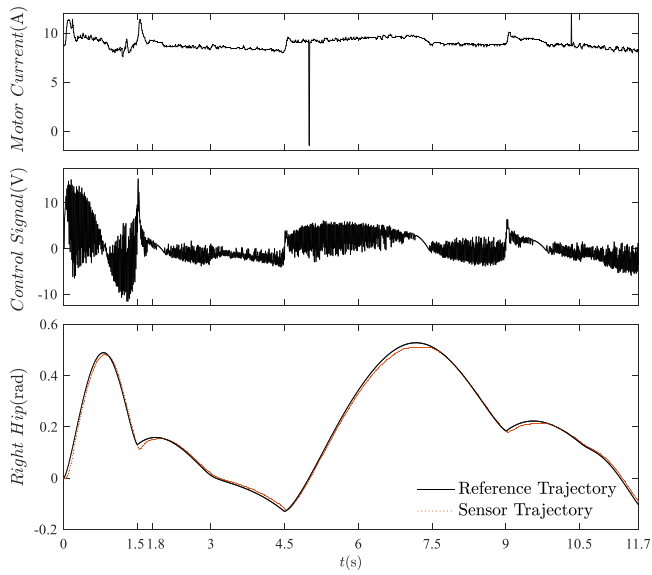

Fig. 24. The reference angle, the real angle, the control signal, and the motor current of the right hip.

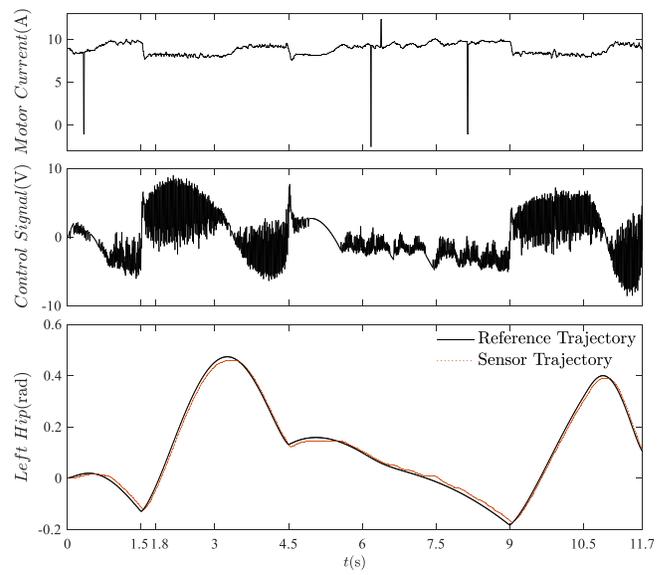

Fig. 26. The reference angle, the real angle, the control signal, and the motor current of the left hip.

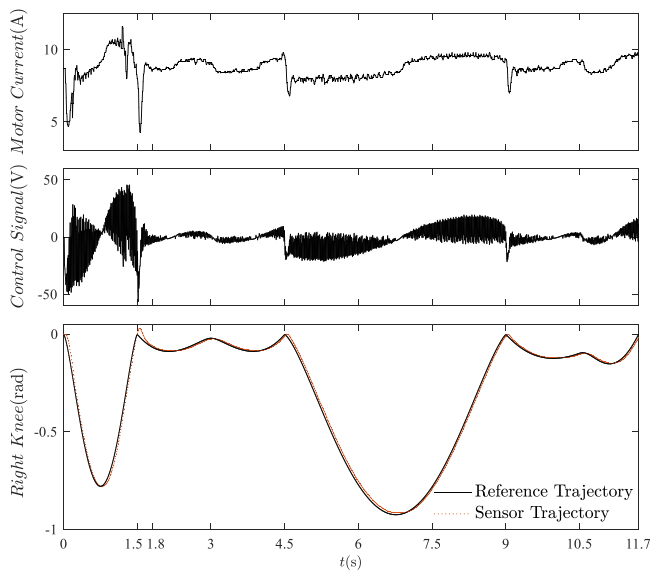

Fig. 25. The reference angle, the real angle, the control signal, and the motor current of the right knee.

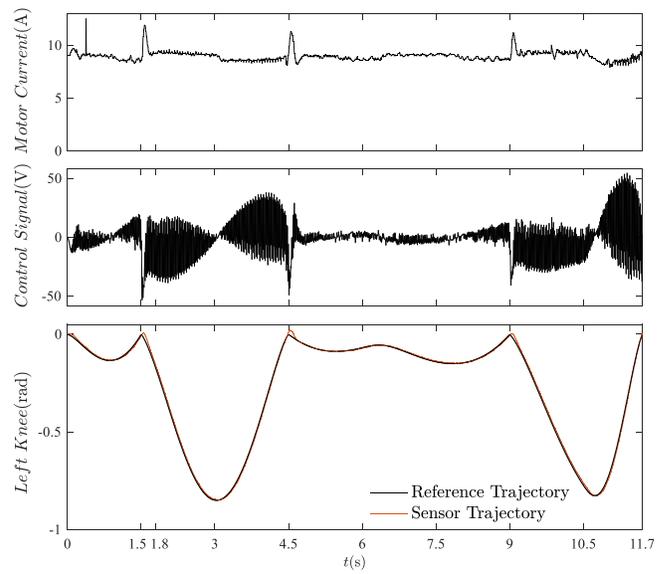

Fig. 27. The reference angle, the real angle, the control signal, and the motor current of the left knee.

## V. CONCLUSION

### REFERENCES

[1] R. S. Mosher, "Handyman to Hardiman," *SAE Trans.*, vol. 76, pp. 588–597, 1968.
[2] A. Formalskii and A. Shneider, "Book Reviews : Biped Locomotion (Dynamics, Stability, Control and Application): by M. Vukobratovich, B. Borovac, D. Surla, and D. Stokich published by Springer-Verlag, 1990," *Int. J. Robot. Res.*, vol. 11, no. 4, pp. 396–396, Aug. 1992.

[3] T. Yan, M. Cempini, C. M. Oddo, and N. Vitiello, "Review of assistive strategies in powered lower-limb orthoses and exoskeletons," *Robot. Auton. Syst.*, vol. 64, pp. 120–136, Feb. 2015.

[4] E. Guizzo and H. Goldstein, "The rise of the body bots [robotic exoskeletons]," *IEEE Spectr.*, vol. 42, no. 10, pp. 50–56, Oct. 2005.

[5] H. Kazerooni, J. L. Racine, L. Huang, and R. Steger, "On the Control of the Berkeley Lower Extremity Exoskeleton (BLEEX)," in *Proceedings of the 2005 IEEE International Conference on Robotics and Automation*, 2005, pp. 4353–4360.

[6] S. Jezernik, G. Colombo, and M. Morari, "Automatic gait-pattern adaptation algorithms for rehabilitation with a 4-DOF robotic orthosis," *IEEE Trans. Robot. Autom.*, vol. 20, no. 3, pp. 574–582, Jun. 2004.

[7] A. Esquenazi, M. Talaty, A. Packel, and M. Saulino, "The ReWalk powered exoskeleton to restore ambulatory function to individuals with thoracic-level motor-complete spinal cord injury," *Am. J. Phys. Med. Rehabil.*, vol. 91, no. 11, pp. 911–921, Nov. 2012.

[8] K. A. Strausser and H. Kazerooni, "The development and testing of a human machine interface for a mobile medical exoskeleton," in *2011 IEEE/RSJ International Conference on Intelligent Robots and Systems*, 2011, pp. 4911–4916.

[9] D. Sanz-Merodio, M. Cestari, J. C. Arevalo, and E. Garcia, "Control Motion Approach of a Lower Limb Orthosis to Reduce Energy Consumption," *Int. J. Adv. Robot. Syst.*, vol. 9, no. 6, p. 232, Dec. 2012.

[10] J. Yoon, R. P. Kumar, and A. Özer, "An adaptive foot device for increased gait and postural stability in lower limb Orthoses and exoskeletons," *Int. J. Control Autom. Syst.*, vol. 9, no. 3, p. 515, Jun. 2011.

[11] R. Huang, H. Cheng, Y. Chen, Q. Chen, X. Lin, and J. Qiu, "Optimisation of Reference Gait Trajectory of a Lower Limb Exoskeleton," *Int. J. Soc. Robot.*, vol. 8, no. 2, pp. 223–235, Apr. 2016.

[12] F. M. Silva and J. A. T. Machado, "Kinematic aspects of robotic biped locomotion systems," in *, Proceedings of the 1997 IEEE/RSJ International Conference on Intelligent Robots and Systems, 1997. IROS '97*, 1997, vol. 1, pp. 266–272 vol.1.

[13] E. T. Esfahani and M. H. Elahinia, "Stable Walking Pattern for an SMA-Actuated Biped," *IEEEASME Trans. Mechatron.*, vol. 12, no. 5, pp. 534–541, Oct. 2007.

[14] T. Kagawa, H. Ishikawa, T. Kato, C. Sung, and Y. Uno, "Optimization-Based Motion Planning in Joint Space for Walking Assistance With Wearable Robot," *IEEE Trans. Robot.*, vol. 31, no. 2, pp. 415–424, Apr. 2015.

[15] D. E. Kirk, *Optimal Control Theory: An Introduction*. Courier Corporation, 2012.